\documentclass[runningheads]{llncs}
\pdfoutput=1
\usepackage[utf8]{inputenc} 
\usepackage[T1]{fontenc}    
\usepackage{hyperref}      
\usepackage{url}         
\usepackage{booktabs}   
\usepackage{amsfonts}    
\usepackage{nicefrac}    
\usepackage{microtype}     
\usepackage{lipsum}
\usepackage{float}
\usepackage{graphicx}
\usepackage{caption, subcaption}
\usepackage{xcolor}
\usepackage{multirow}
\usepackage{amsmath}
\newenvironment{myproof}{ 
\noindent {\it Proof} \noindent} {\hfill $\Box$\vskip 5mm} 

\begin{document}

\title{  Closed pattern mining of interval data and distributional data}
\author{
   Henry Soldano \inst{1,2,3}
 Guillaume Santini \inst{2}
 \and
 Stella Zevio \inst{2}
%
}

\institute{ NukkAI, Paris, France \and LIPN CNRS UMR 7030, Université Sorbonne Paris Nord, Villetaneuse, France \and
Muséum d'Histoire Naturelle, ISYEB, Paris, France\\
}

\maketitle

\begin{abstract}

We discuss pattern languages for closed pattern mining and learning of interval data and distributional data. 
We first introduce  pattern languages relying on pairs of intersection-based constraints or pairs of inclusion based constraints, or both, applied to intervals. We discuss the encoding of such interval patterns as itemsets thus allowing to use closed itemsets mining and formal concept analysis  programs. We experiment these languages on clustering and supervised learning tasks.
Then we show how to extend the approach to address distributional data.

\end{abstract}

\section{Introduction}
We investigate how to address interval and distributional data for mining and learning purpose.
   In pattern mining   a variable value may happen to be  an interval or a distribution. This may be by nature, the interval representing some time period, or because the object to represent is  a group, as a specie, or because there is some uncertainty on the measure of a numerical value.  In  the bibliographical data  that motivated the present work, we had to consider   authors and had to represent their period of publication, i.e. some time interval (in years). Previous  work about extending  mining  and learning techniques to interval  data includes  those from \cite{Bock:2000aa,Gioia:2006aa,Cabanes:2013aa}.

    A pattern constraining a numerical variable $Y$  is typically of the form $ y \in M$ where $M$ is some interval.
    Such patterns  are obtained either   indirectly\cite{Kaytoue:2011aa} or directly through  \emph{interordinal scaling}\cite{Ganter:1999aa}, i.e. associating to $y$  a finite set of half-bounded interval  constraints $ y \in M_i$.
     Now, to  occur in an object subset $S$ a pattern  has to satisfy  $\Delta \subseteq M$ where  the interval $\Delta$ contains all $y$ values of objects in $S$. Such  \emph{inclusion based patterns}   directly  applies to interval data \cite{Kaytoue:2010aa}. 


Our first  purpose  is to investigate pattern languages on  interval data. For that purpose, aside from inclusion based patterns, we introduce  \emph{intersection based} patterns, i.e. atomic patterns obtained by considering  intersection of the interval value with semi-intervals, i.e. of the form  $\Delta \cap M \neq \emptyset$. 

Our second purpose is to address distributional data that  expresses  the uncertainty about  the value of some  numerical    variable in an object  as a \emph{cumulative distribution function}(a \emph{cdf}) \cite{Kriegel:2005tf,Chau:2006vr,Gullo:2008aa,Bock:2000aa}. 
We  investigate  how to address distributional data by translating a distribution into one or more intervals.



After a light presentation of closed pattern mining (CPM) in Section \ref{abstractPatternMining}, we  define and investigate   the intersection-based pattern language $L_\mathit{I}$ in Section \ref{LII} 
 and discuss its  patterns encoding 
 as  itemsets, thus allowing use of standard CPM and Formal Concept Analysis (FCA) tools. We then illustrate the resulting closed patterns in our motivating application. In Section \ref{LCLIILIIC} we consider  inclusion-based pattern language $L_C$ and  discuss $L_\mathit{IC}$ that integrates both inclusion-based and intersection-based patterns.  In Section\ref{experiments} we experiment $L_{C}$, $L_\mathit{I}$ and $L_\mathit{IC}$ on clustering and supervised learning tasks. Section \ref{distributionData} define patterns for  distributional data and provide some further experiments.



 \section{Formal Concept Analysis and Closed Pattern Mining}\label{abstractPatternMining}
 \label{sec:bipattern}
 
 
We present necessary results and definitions in their   closed pattern mining formulation  (see \cite{Soldano:2019aa}), however the translation to  the FCA \emph{pattern structure} formulation is straightforward (see \cite{Kaytoue:2011aa} for its use in numerical pattern mining). 

 A pattern $q$ has an \emph{extension} also called a \emph{support set} $e=\mathrm{ext}(q)$ representing its set of occurrences in a set of objects $V$.  When considering the  equivalence class  of all patterns with support set $e$, we have the following result:  
\begin{proposition}\label{propFond}
Let $L$ be a pattern language partially  ordered by \emph{specificity}, a partial order such  that $q'\geq q$ implies that $ \mathrm{ext}(q')\subseteq   \mathrm{ext}(q)$.Assume that $(L, \leq)$ is a lattice and  that for any object $o$ there exists a unique most specific pattern $d(o)$ that  occurs in $o$. Let  $q$ be a pattern, then  
\begin{itemize}
\item the  class of patterns with same support set as $q$ has a greatest element $f(q)$
\item $f$ is a closure operator
\end{itemize}
\end{proposition}

 A closed pattern is obtained by using an intersection operator $\mathit{int}$ that applies the lowest upper bound operator $\land$ to a set of object descriptions $d[e]$\footnotemark. 
 \footnotetext{$d[e]$ is the image of $e$ by $d$, i.e. $d[e]=\{d(v) | v \in e\}$}
 The most specific pattern $f(q)$ of the class of pattern with   support set $e=\mathrm{ext}(q)$ is: 
\begin{eqnarray} \label{three}
f(q)& =&\mathrm{int} \circ \mathrm{ext}(q) \mbox{ where }\\ 
\mathrm{int}(e)&=& \bigwedge_{o \in e} d(o)
\end{eqnarray} 

In standard FCA and   itemsets CPM, objects are described as itemsets i.e. subsets of a set of items $I$. In this case the intersection $\land$ operator simply is the set theoretic intersection operator $\cap$, i.e. $\mathrm{int}(e)= \bigcap_{o \in e} d(o)$ which represents the set of items belonging to all objects in $e$. %

FCA is focussed on the partial ordering of such closed patterns. The set of pairs $(c,e)$, also called \emph{concepts},  where $c$ is a closed pattern and $e$ its support set $e=\mathrm{ext}(c)$, form a lattice called a \emph{concept lattice}. 
\section{$L_\mathit{I}$ interval patterns and their encoding}\label{LII}

\label{sec:results}

\subsection{Informal presentation of inclusion-based and Intersection-based Interval patterns}\label{informalPresentation}
In this section we consider a variable $\Delta$ whose values are integer intervals $i .. j$ included in the domain $D=1..5$.
We simply denote such an  interval by a digit word, e.g. $12$  stands  for $1..2$.   We also consider the set of  left-bounded semi-intervals  ${\cal M}_L$  and the set of right-bounded  intervals ${\cal M}_R$ (including the whole domain $D$ in both of them), i.e.  ${\cal M}_R= \{1, 12, 123, 1234,12345\}$ and 
${\cal M}_L = \{5, 45, 345, 2345,12345\}$.
We then consider objects $a$ and $b$ with $\Delta$ values  $\Delta(a)=12, \Delta(b)=234$ and 
consider 
 pattern languages on such interval variables. 

\subsubsection{Inclusion-based interval  patterns} have the form $\Delta \subseteq M$.
 For instance,   if the  variable $\Delta$  has value   $\Delta(b)=234$ in object $b$,  we  may state that pattern $\Delta \subseteq  1234$ occurs in $b$ while pattern $\Delta \subseteq  123$  does not. 
 These patterns are the elements of the pattern language $L_C$.
 Note that any   $\Delta \subseteq l .. r$  interval pattern is the conjunction of  two \emph{half-bounded} interval patterns  $(\Delta \subseteq M_L,\Delta \subseteq M_R)$ where $M_L\in {\cal M}_L$  and $M_R\in {\cal M}_R$.  For instance, in our illustrative example  
 we may  write  pattern $\Delta \subseteq 234$  as 
 $(\Delta \subseteq 2345, \Delta \subseteq 1234)$. 


\subsubsection{Intersection-based interval patterns}
are  new constraints on interval values  of the form  $\Delta \cap M\not =\emptyset $, we simply denote by  $\Delta \cap M $, where $M \in {\cal M}_R$ or  $M \in {\cal M}_L$.  Such a constraint holds  whenever $\Delta$ intersects  $M$.
 We then define  a pattern in the pattern language $L_\mathit{I}$ as the conjunction of  a left-bounded constraint and a right-bounded  constraint. For instance $(\Delta \cap 2345,\Delta \cap 1234)$ is a pattern of  $L_\mathit{I}$. 
 
 Finally note that  regarding intersection-based patterns as well as inclusion-based pattern, both left-bounded constraints and right-bounded constraints are totally ordered, e.g.  $\Delta \subseteq 12$  implies  $\Delta  \subseteq 1234$ and  $\Delta \cap 12$  implies  $\Delta  \cap 1234$.
 \subsubsection{Comparing intersection-based to inclusion-based patterns}
%
 We represent  Table  \ref{tab:intervalValues}  the interval values for  objects $a$ and $b$, and for both pattern languages, their  descriptions as well  as the most specific pattern occurring in both $a$ and $b$.   

   

In $L_\mathit{I}$,  $\Delta(a)=12$ is  described  as  $d_\mathit{I}(a)=$ $(\Delta \cap 2345,\Delta \cap 1)$ as i) $\Delta \cap 2345$ is the strongest left-bounded constraint satisfied by $\Delta(a)=12$ ($2345$ is the smallest left-bounded interval that intersects $12$)
and ii) $\Delta \cap 1$ is the strongest right-bounded constraint satisfied by $\Delta(a)$.
In the same way, $\Delta(b)=234$ is described as  $d_\mathit{I}(b)=$ $(\Delta \cap 45,\Delta \cap 12)$.

Now the most specific  $L_\mathit{I}$ pattern $\mathrm{int}_\mathit{I}(\{a,b\})$ occurring in both $a$ and $b$   is  obtained by considering separately left and right half-bounded constraints and finding in each case the weakest among the  constraints in objects $a$ and $b$,  i.e. $\mathrm{int}_\mathit{I}(\{a,b\})=(\Delta \cap 2345,\Delta \cap 12)$.

Following the same process,  $\Delta(a)=12$ is described in  $L_\mathit{C}$ by  $(\Delta \subseteq 12345,\Delta \subseteq 12)$) and $\Delta(b)=234$ is described  by  $(\Delta \subseteq 2345, \Delta \subseteq 1234)$.  The most specific $L_\mathit{C}$ pattern  occurring in  $\{a,b\}$ is 
$\mathrm{int}_C(\{a,b\})=(\Delta \subseteq 12345,\Delta \subseteq 1234)$. 
 \begin{table}[H]  

    \centering
    {\footnotesize
      \caption{Most specific   $L_C$ and $L_\mathit{I}$  patterns occurring in object subsets.    \label{tab:intervalValues} 
    }

    \begin{tabular}{|c|cccccc|cc|c||cc|c|}
    \hline
     $S$	& $\Delta$ &	&	&	&	&	&$\mathrm{int}_\mathit{I}(S)$	&						& $\equiv_\mathit{I}$						& $\mathrm{int}_\mathit{C}(S)$		& 						&$\equiv_C$  \\
    \hline 
    $\{a\}$	& 1		& 2	&	& 	&	&	&$\Delta \cap 2345$, 			& $\Delta \cap 1$ 		&$\Delta \supseteq  12$				&	$\Delta \subseteq 12345$, 		& 	$\Delta \subseteq 12$	&		$\Delta \subseteq 12$\\
   $\{b\}$	& 		& 2	& 3	&4	&	&	&$\Delta \cap 45$,				& $\Delta \cap 12$  		&$\Delta \supseteq 234$				&	$\Delta \subseteq 2345$,			&	$\Delta \subseteq 1234$	&	 	$\Delta \subseteq 234$		\\
\hline
 $ \{a,b\} $& 		& 	& 	&	&	&	&$\Delta \cap 2345$,				 &	$\Delta \cap 12$	&$\Delta \cap 2$ 					& 	$\Delta \subseteq 12345$,			&	$\Delta \subseteq 1234$ 	&		$\Delta \subseteq 1234$\\
 \hline
           \end{tabular}
    }
 \end{table}

\subsubsection{Interpreting patterns}
Recall  that  any $L_C$ pattern  rewrites as  $\Delta \subseteq M_L \cap M_R$
We call this  simpler writing the \emph{interpretation} of the pattern in $L_\mathit{C}$ (see column $\equiv_C$ in Table \ref{tab:intervalValues}). 
 Interpretation is less straightforward  regarding  $L_\mathit{I}$ patterns (see  column $\equiv_\mathit{I}$ in Table \ref{tab:intervalValues}). 
Let us consider  pattern $(\Delta \cap M_L, \Delta \cap M_R)$:
\begin{itemize}
\item  Whenever $M_L \cap M_R$ intersects  the pattern  rewrites as  $\Delta \cap (M_L \cap M_R)$, for instance $\mathrm{int}_\mathit{I}(\{a,b\})=(\Delta \cap 2345,\Delta \cap 12$) rewrites as $\Delta \cap 2$. 
\item Whenever $M_L$ and $M_R$  does not intersect,  $\Delta$ has to include both the  minimal value of $M_L$  and the maximal value of  $M_R$  thus resulting in a $\Delta \supseteq M$ interpretation. For instance $d_\mathit{I}(b)=$ $(\Delta \cap 45, \Delta \cap 12)$ rewrites as $\Delta \supseteq 234$.
\end{itemize}


In the next section,
  we give a formal and more  general presentation of $L_\mathit{I}$.
\subsection{Interval patterns in $ L_{\mathit{I}} $ }
We  consider  objects $o$ described by an interval value  $\Delta(o)$ included in some domain $D$ that  we consider, with no generality loss,  to be of the form $]\delta_m,\delta_M]$. 

$L_{\mathit{I}}^k$ relies on  half-bounded intervals whose bounds  belongs to 
 $T=\{s_1, \dots ,s_k\}$:

\begin{itemize}
\item $\Delta \ \cap \ ]s_i, \delta_M]$,  we also write   $\cap >s_i$, occurs in $o$   when $\Delta(o)$ intersects $]s_i, \delta_M]$.
\item $\Delta \ \cap \ ]\delta_m,s_i]$, we also write $\cap  \leq s_i$ occurs in $o$  when $\Delta(o)$ intersects $]\delta_m,s_i]$.
\end{itemize}
 By denoting $\delta_m$ by  $s_0$ and $\delta_M$ by $s_{k+1}$ we add two (virtual) constraints  $\cap >s_0$ and $\cap \leq s_{k+1}$ which are always true.  We may  then   define  a $L_{\mathit{I}}^k$ pattern \footnotemark as the conjunction  $(\cap >s_i, \cap \leq s_j)$ with $i \in 0.. k$ and $j \in 1.. {k+1}$.
 We denote by $T^+$ the bounds set $T \cup \{s_0, s_{k+1} \}$. 
If no confusion is possible  we  write $L_{\mathit{I}}^k$ as $L_{\mathit{I}}$. 
 
 \footnotetext{$\mathit{I}$ stands for Intersects}

Patterns  are partially ordered by \emph{specificity} and  $q_1$ is more specific  than $q_2$, i.e. $q_1 \geq q_2$, whenever any object interval $\Delta$  that satisfies $q_1$ also satisfies $q_2$. 
 The specificity order defines $ L_{\mathit{I}} $ as a lattice which in the finite case   only requires that any  pattern pair $q_1,q_2$  have a unique greatest lower bound $q_1 \land  q_2$ in $L_\mathit{I}$. Namely, $q_1 \land  q_2$ is the unique most specific  pattern such that 
  $q_1 \land  q_2 \leq q_1$ and  $q_1 \land  q_2 \leq q_2$:
 
 \begin{proposition}\label{LIIorder}
 Let $q_1 =(\cap >s_{i_1}, \cap \leq s_{j_1})$ and $q_2=(\cap>s_{i_2},\cap  \leq s_{j_2})$ be two patterns in $L_\mathit{I}$ with $s_{i_1}, s_{j_1}, s_{i_2}, s_{j_2} \in T^+$ then 
 \begin{eqnarray}
 q_1 \geq q_2 & \mbox{ iff } & s_{i1}  \geq s_{i_2}  \mbox{ and } s_{j_1} \leq s_{j_2}\\
 q_1 \land q_2 &=&(\cap >s_{i}, \cap \leq s_{j}) \mbox{ where }\\
 s_i &=&\mathrm{min}(s_{i_1},s_{i_2})  \mbox { and }\\
  s_j &=&\mathrm{max}(s_{j_1},s_{j_2})
   \end{eqnarray} 
 \end{proposition}
 For instance $\cap > s_i$ is more specific than $ \cap >s_{i-1}$  while $\cap \leq$$s_i$ is more specific than  $\cap  \leq$$s_{i+1}$. 
As a consequence  $(\cap >s_2,\cap \leq s_4)$ 
is more specific than  
 $(\cap >s_1, \cap \leq s_5)$  but is not more specific than $(\cap >s_1,\cap \leq s_3)$.


 \subsection{Interpreting $L_\mathit{I}$ patterns}\label{interpretations}
 
A pattern as  $(\cap>s_1, \cap\leq s_5)$ occurs  $\Delta$ intersects $]s_1,s_5]$, i.e. $\Delta \cap ]s_1,s_5]$  while a  pattern as $(\cap>s_5, \cap \leq s_2)$ may be   interpreted as $\Delta \supseteq  [s_2, s_5+\epsilon]$ when   intervals $\Delta$ are defined with some resolution $\epsilon$, for instance $\epsilon=1$ when bounds are integers. 
 To summarize we may interpret  $L_{\mathit{I}}$ as made of two kind of patterns:
 
  \begin{proposition}[Interpreting $L_\mathit{I}$ patterns]
  In $L_\mathit{I}$ 
   \begin{eqnarray}
\Delta \ \cap \ ]s_i,s_{i+m}] \mbox{ interprets }  (>s_i, \leq s_{i+m})\mbox{ whenever }m>0 \label{intersectsInterpretation}\\
\Delta \ \supseteq  [s_{i-m}, s_{i}+\epsilon]  \mbox{  interprets  } (>s_i \leq s_{i-m}) \mbox{ whenever } m\geq 0 \label{supersetInterpretation}
\end{eqnarray}

\end{proposition}
\subsection{$L_\mathit{I}$ and the concept lattice of indistinguishable objects}
We consider that  two interval objects $\Delta$ are \emph{indistinguishable} whenever they occur in exactly the same patterns. Indistinguishability is an equivalence relation and  we  consider   the set $O^k$ of its equivalence classes. We have then:

\begin{proposition}
The  set  $O^k$ of distinguishable object intervals  in $L_\mathit{I}^k$ has size
\begin{equation}
\mid O^k\mid = (k+1)(k+2) / 2
\end{equation}

\begin{example}\label{exS1S2}
 Figure \ref{Fig-distinguishable6} represents the set $O^2$ of 6 distinguishable intervals representing objects using $L^2_\mathit{I}$ patterns: only interval object pairs $\Delta, \Delta^\prime$ that are included in different $]s_i,s_{i+1}]$ intervals or cross a  different set of frontiers $s_i$ are distinguishable in  $L^2_\mathit{I}$ : 3 cross no $s_i$ frontier, 2 cross 1 frontier and 1 cross 2 frontiers leading to  $\sum_{f=0}^{f=k} k+1-f =\sum_{f=0}^{f=k+1} f = 3*4/2=6$ interval objects.
  \begin{figure}[h]
    \centering
    \includegraphics[width=0.6\textwidth]{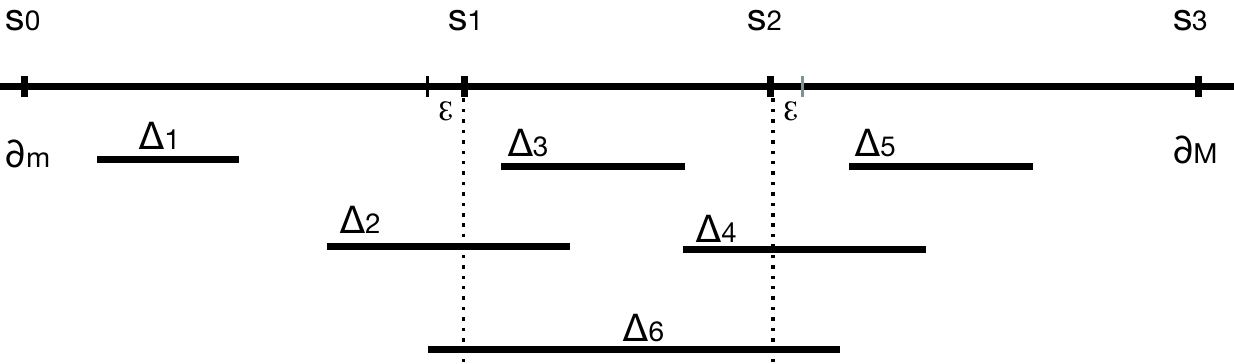}
    \caption{The set of 6 distinguishable interval objects when $T=\{s_1,s_2\}$
    }
  \label{Fig-distinguishable6}
\end{figure}
\begin{figure}[h]
    \centering

      \includegraphics[width=0.95\textwidth]{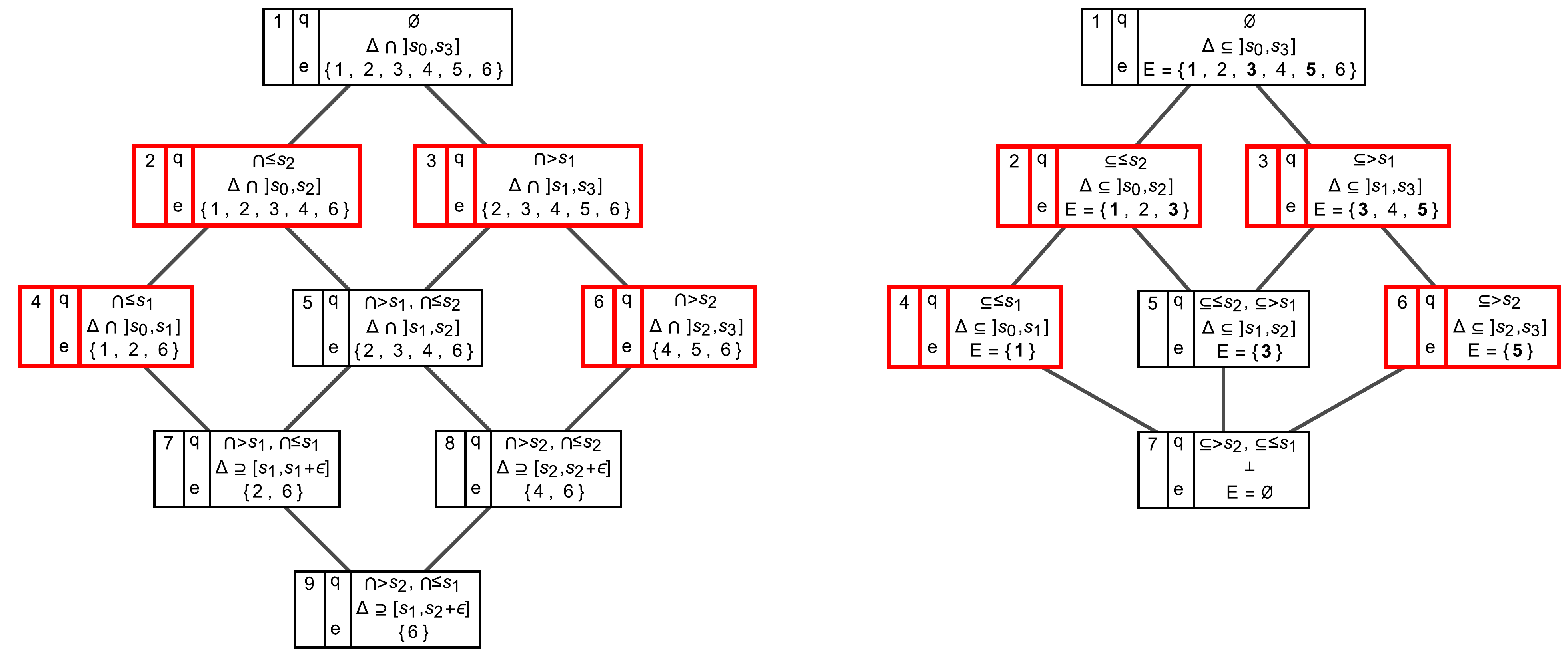}
    \caption{The lattices $L^2_\mathit{I}$   and  $L^2_\mathit{C}$  defined on $T=\{s_1,s_2\}$ and represented as  concept lattices with respect to $O_2$. Each pattern is displayed as $\cap >s_i,\cap \leq s_j$  (omitting the  $\cap >s_0$ and $\cap \leq s_3 $ trivial constraints) together  with its interpretation and  its occurrences in $O^2$. $\lor$-irreducible elements are in red rectangles.
    }
  \label{Fig-IIetLC}
\end{figure}
\end{example}

\end{proposition}
\begin{example}\label{exLatticeLII}
 Figure \ref{Fig-IIetLC}-left) displays the lattice ($L_\mathit{I}^2, \leq)$defined on $T=\{s_1,s_2\} $. On top we have  the least specific pattern  $\Delta \ \cap \ ]s_0,s_3[$,  which occurs in any interval $\Delta$,    and on bottom the most specific pattern $\Delta \ \supseteq  [s_1, s_2+\epsilon]$.  
%

\end{example}

For sake of simplicity, we have excluded the  $\emptyset$  interval from the $\Delta$ domain.
 That 
 would result in  an additional  object in $O^k$    and a new bottom node $\Delta=\bot$ in  $L_\mathit{I}$.
%
 
  \subsection{Closed pattern mining on $L_\mathit{I}$ }
\subsubsection{Object  description }

 Proposition  \ref{propFond} assumes 
 that  any object  $o$ as a  unique most specific pattern $d(o)$ among patterns that occurs  in $o$, which is ensured by: 
 \begin{proposition}
Let $s_l$ and $s_u$ be defined as follows:
 \begin{itemize}
 \item $s_{l}$ is the greatest $s_i$ in $T^+$ such that $\cap>s_i$ holds on  $\Delta(o)$
  \item $s_{u}$ is the smallest $s_j$ in $T^+$  such that $\cap\leq s_j$ holds on $\Delta(o)$
 \end{itemize}

The most specific  $L_\mathit{I}$ pattern that occurs in   $o$ is  
 $  d(o)= (\cap>s_{l},\cap \leq s_{u})$ 

%
 \end{proposition}
 \begin{myproof}
 First, by hypothesis  $(>s_{l}, \leq s_{u})$ occurs in $o$ as both constraints are satisfied. Second, patterns $(>s_i,\leq s_j)$ with either $i>l$ or $j< u$, also by hypothesis, does not occur in $o$.  Finally  all patterns    $(>s_i,\leq s_j)$  with $i \leq l$ and $ j \geq  u$ are less specific than  $(>s_{l},\leq s_{u})$ (the constraints are weaker). As a consequence $(>s_{l},\leq s_{u})$ is the most specific pattern that occurs in $o$.
 Furthermore,  assume $u > l+1$, it  means that $\Delta(o)$ intersects $]s_{l}, s_{u}]$, but in this case $\Delta(o)$ also intersects  $]s_{l}, s_{i_{l+1}}]$ and therefore $(>s_{l}, \leq s_{l+1}])$  occurs in $o$. This means that   $(>s_{l}, \leq s_{u})$ is  not the most specific pattern occurring in $o$. Therefore we have $   u \leq l+1$.
  \end{myproof}
\begin{example}\label{exo2o4-1}

Consider objects $o_2$ ad $o_5$  with  interval objects $\Delta_2$ and $\Delta_5$ from $O^2$ displayed Figure \ref{Fig-distinguishable6} together with  the $L^2_\mathit{I}$ lattice displayed Figure \ref{Fig-IIetLC}-left. We have:   
	$d(o_2)=(\cap>s_1, \cap\leq s_1) \mbox{ i.e. } \Delta \supseteq [s_1, s_1+\epsilon] \mbox{ (node 7 Figure \ref{Fig-IIetLC}-left) }$
	
	$d(o_5)=(\cap >s_2,  \cap\leq s_3)\mbox{ i.e. } \Delta \cap [s_2, s_3] \mbox{ (node 6 Figure \ref{Fig-IIetLC}-left) }$\\
\end{example}
%
%
%
%
%
 \subsubsection{Closed intersection based pattern mining}
 As $L_\mathit{I}$ is a lattice and that each object as a unique description in $L_\mathit{I}$  Proposition \ref{propFond} states that closed patterns with respect to an object set are properly defined. The  closed pattern with same   occurrences as any pattern $q$  is then defined in Equation \ref{three}. 
 \begin{example}
  We consider  $L^2_\mathit{I}$ displayed Figure \ref{Fig-IIetLC}-left together with an object set $\{o_2,o_3,o_4,o_5\}$ with respective $\Delta$ values  $\{\Delta_2,\Delta_3,\Delta_4,\Delta_5\}$ extracted from $O^2$. The top pattern $\top$  (node 1) has extension $\mathrm{ext}(\top)=\{o_2,o_3,o_4,o_5\}$ whose objects are respectively described in nodes $7, 5, 8, 6$ (the object description is the first node from bottom to top in which extension the object appears). We have then that the  closed pattern $(\cap>s_0,\cap \leq s_3)$ with same extension as the top pattern  $\mathrm{int}(\{o_2,o_3,o_4,o_5\}$ is found in their lowest ancestor in the lattice, namely node 3. 
%
%
%
 \end{example}

\subsection{Intersection based pattern encoding as an itemset}\label{encoding}

Encoding of  $L^k_\mathit{I}$ patterns as itemsets allows using standard FCA and itemsets CPM  algorithms to enumerate closed patterns. 
%
As $L_\mathit{I}^k$ is a lattice, any pattern $q$ has a maximal  representation as the subset of $\lor$-irreducible\footnotemark elements less specific than $q$.
\footnotetext{A $\lor$-irreducible element $q$ cannot be obtained as a the join $q' \lor q''$ of two elements $q'$ and $q''$. In a powerset $2^I$  they are the singletons.}
  These irreducible elements are the  $\cap>$$s_i$ and  $ \cap\leq$$s_i$  patterns.  Hence the following proposition:
\begin{proposition}\label{forth} Let $q=(\cap >s_i, \cap \leq s_j)$  be a pattern in $L_\mathit{I}$ then

$q$ rewrites as $R(q)=\{\cap >s_{v} \mid v \  \leq i \in T \} \  \cup \ \{\cap  \leq s_{v} \mid v  \ \geq j \in T\}$
 
\end{proposition}


Conversely, the original form $q=(\cap>s_i, \cap \leq s_j)$ of any pattern   may be retrieved from this maximal representation:

\begin{proposition}\label{back}
Let $R(q)$ be the maximal representation of  pattern $q$, and let 
\begin{itemize} 
\item $\mathit{i_L}$ be the greatest $i$ such that $\cap >s_i $ belongs to $R(q)$   if it exists
\item $\mathit{i_R}$ be  the smallest  $i$  such that $\cap \leq s_i $ belongs to $R(q)$ if it exists
\end{itemize}
Then we have  $q=(\cap>$$s_\mathit{i_L}$, $\cap\leq$$s_\mathit{i_R})$ 
\end{proposition}

\begin{example}
In
$L^2_\mathit{I}$ displayed in Figure  \ref{Fig-IIetLC}-left,
the  $L_\mathit{I}$ pattern  $ (\cap>$$s_2$,$\cap\leq$$s_2)$ (node 8)  has maximal representation $\{\cap>s_1$,$\cap>s_2$,$\cap\leq s_2\}$
, figuring  $\lor$-irreducible elements  above $q$ in the lattice.  
\end{example}



%
Using  maximal representations, we may 
   implement  the $ \mathit{int}$  and $ \mathit{ext}$ operators:
    \begin{proposition}\label{maximalRepresentationLII}
Let $q_1$ and  $q_2$  be $L_\mathit{I}$ patterns, then
\begin{eqnarray}
 q_1 \geq q_2 &\mbox{ if and only if }& R(q_1) \supseteq R(q_2)\\
R(q_1 \land q_2) &=&R(q_1) \cap  R(q_2) \\
 \mathrm{ext}(q) &=&\{ o \mid R(q) \subseteq R(d(o))\}  \\
\mathrm{int}(\{o_1, \dots o_n\}) &=& \bigcap_{i \in \{1 \dots n\}} R(d(o_i))
\end{eqnarray}
\end{proposition}

%

\subsection{Using $L_\mathit{I}$ to mine bibliographic data}\label{bibliographic}
In the  experiments discussed in this section, as well as experiments discussed  in Sections \ref{experiments} and \ref{distributionalData} closed pattern enumeration is performed using the top-down closed itemset mining  program from the minerLC  software\footnotemark  \cite{Soldano:2019aa}.  
\footnotetext{\url{https://lipn.univ-paris13.fr/MinerLC/}}
  
  Table \ref{tab:12-patterns}   displays a pattern set selection
  made of 12 closed bi-patterns
  \footnotemark   from a study regarding expert retrieval from semantic annotation of authors and cited publications. 
 \footnotetext{When mining  bi-partite graphs built on two vertex set $V_1$ and $V_2$, a  bi-pattern is a pair of patterns applying respectively to objects from $V_1$ and from $V_2$.}
   A bi-pattern displays constraints about authors and about publications cited by authors in a bipartite graph.
   Only authors have a publication period $\Delta_A$ while   publications have a publication year $Y_P$.  Interval data regarding the  publication period $\Delta_A$ of authors is handled using a
   $L_\mathit{I}^4$ pattern language.
 The thresholds in $T$ have been defined as $\{ 1992,1999,2004,2007\}$ with bounds $s_0=1985$ and $ s_5= 2012$ and a resolution $\epsilon=1$ year. 
 In order to ease the reading we  rewrite the interval subpatterns using closed intervals, i.e. for instance $\Delta_A \cap ]2007,2012]$ is rewritten as   $\Delta_A\cap [2008,2012]$.
 
 The $\Delta_A$ subpatterns  range from very general to very specific ones.  For instance, $\Delta_A \ \cap\  [1993,2007]$ appearing in  bi-pattern $2$ only requires an author to have published some  article  between 1993 and 2007, while   $\Delta_A \ \supseteq\ [1999,2000]$ from bi-pattern $11$ requires the publication period of the author to include
 $[1999..2000]$.

%

\begin{table}[H]
    \centering
    {\footnotesize
    \begin{tabular}{ll}
    P & Description \\
    \hline
   1 &  inf.\_extr., $\Delta_A \ \cap\  [2008,2012]$, $Y_P > 1999$ \\
    2 & inf.\_extr., nat.\_lang.\_processing, $\Delta_A \ \cap \  [1993 ,2007] $, $Y_P \leq 2007$ \\
    3 & inf.\_extr. $\Delta_A \  \supseteq \ [1992,1993]$, $Y_P \leq 2004$\\
    4 & inf.\_extr., languages.., $\Delta_A\ \supseteq\  [2004,2008]$, $Y_P \leq 1999$ \\
    5 & inf.\_extr., user\_information, $\Delta_A \ \cap\  [1993 ,1999]$, $1992 < Y_P \leq 2007$\\
    6 & inf.\_extr., learning, $\Delta_A \ \supseteq\ [2004,2005]$, $1992 < Y_P \leq 2007$ \\
    7 & inf.\_extr., $\Delta_A \ \supseteq\ [1999,2005]$, $2004 < Y_P \leq 2007$\\
    8 & inf.\_extr., cond.\_random\_field, $\Delta_A \ \cap\ [2005,2007]$, $1999 <Y_P \leq 2007$\\
    9 & inf.\_extr., languages, named\_entity\_recog., $\Delta_A \  \supseteq \ [2004,2008]$,
     $1992 < Y_P \leq 2007$\\
    10 & inf.\_extr., correl.\_analysis,  $\Delta_A \  \supseteq \ [1999,2008]$, $Y_P \leq 1999$\\
   11 & inf.\_extr., languages, nat.\_lang.\_processing, $\Delta_A \ \supseteq\ [1999,2000]$,
       $1992 < Y_P \leq 1999$\\
   12 & inf.\_extr., named\_entity\_recognition,  $\Delta_A \ \cap \  [1993 ,2004] $,
   $1999 < Y_P \leq 2004$
    \end{tabular}
    }
    \caption{12  bi-patterns from the semantic  annotation problem.
    }
    \label{tab:12-patterns}
\end{table}

\section{$L_\mathit{I}$, $L_\mathit{C}$ and $L_\mathit{IC}$}\label{LCLIILIIC}


$L_C$    
is a variant of interordinal scaling  \cite{Ganter:1999aa}  
   which as noticed in \cite{Kaytoue:2011ab}  leads to a pattern language for  interval variables. 
   $L_C^k$ is built on    the following constraints:
\begin{itemize}
\item $\Delta \ \subseteq \ ]s_i, s_{k+1}]$ further referred to as $ \subseteq >s_i$ and
\item  $\Delta \ \subseteq \ ]$$s_0,s_i]$ further referred to as $ \subseteq \leq s_i$
\end{itemize}
In the same way as in $L_\mathit{I}$ a pattern in $L_\mathit{C}$ is of the form $(\subseteq >s_i, \subseteq \leq s_j) $ with $i,j \in T^+$ and $i < j$ to which is added the always false pattern $\bot$. We refer to  \cite{Kaytoue:2011ab} for  closed pattern mining in such a language. The lattice is defined in a very similar way as for $L_\mathit{I}$ and indeed  Proposition \ref{LIIorder}, when replacing within patterns $\cap$ by $\subseteq$, as well as  Proposition \ref{maximalRepresentationLII}, still hold.  $ L_\mathit{C} $ is different from  $ L_\mathit{I} $ 
 as we can see by comparing   $L^2_{\mathit{I}}$ (Figure \ref{Fig-IIetLC}-left) and  $L^2_\mathrm{C}$ (Figure \ref{Fig-IIetLC}-right) on indistinguishable objects from $O^2$ (see Figure \ref{Fig-distinguishable6}). 
Interpretations are obtained as follows:

  \begin{proposition}
 In  $L_\mathit{C}$ $\Delta \ \subseteq \ ]s_i,s_{i+m}] $ $\mbox{ interprets } $ $ (\subseteq >s_i, \subseteq \leq s_{i+m})\mbox{ for }m>0 $
\end{proposition}

\subsection{ $L_\mathit{IC}$} 

\begin{figure}[h]
    \centering
   \includegraphics[width=0.8\textwidth]{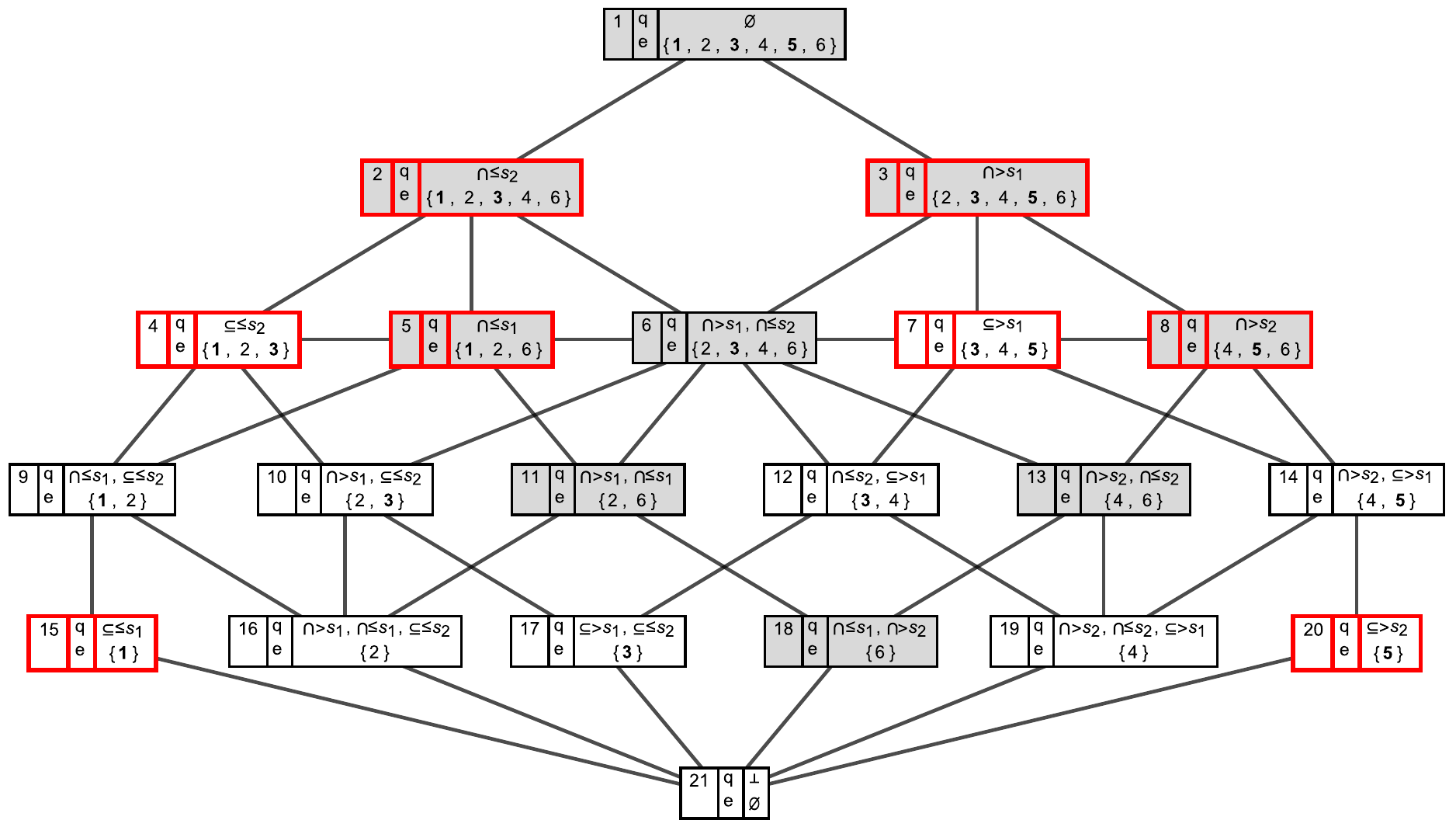} 
       \caption{The lattice $L^2_\mathit{IC}$  defined on $T=\{s_1,s_2\}$. Nodes belonging to  $L^2_\mathit{I}$ are grey-colored while $\lor$-irreducible constraints 
       are within red rectangles.
    }
  \label{figLIII-s1s2}
\end{figure}

 We consider now the language $L^k_\mathit{IC}$ that mixes constraints from  $L^k_\mathit{I}$  and from 
 $L^k_\mathit{C}$. Note that   constraints from $L^k_\mathit{C}$ are negations of  constraints from $L^k_\mathit{I}$:  $\Delta \subseteq ]s_i, s_{k+1}]$ holds whenever $ \Delta \cap ]s_0, s_i]$ fails and $\Delta \subseteq ]s_0, s_i]$ holds whenever $ \Delta \cap ]s_i, s_{k+1}]$ fails.
In $L_\mathit{IC}^2$ displayed Figure \ref{figLIII-s1s2} the nodes from $L_\mathit{I}^2$ are represented as grey-colored rectangles while $\lor$-irreducible patterns, i.e. atomic constraints from both languages, are in red rectangles. Note that  some nodes are from neither  $L_\mathit{I}^2$  nor $L_\mathit{C}^2$ as node 19 (with extension $\{2\}$ in $O^2$) and node 16 (with extension $\{4\})$. 
%

%
%
%

\footnotetext{standing for Intersects Includes and Contained} 

\begin{example}\label{exLIIC1}
The $L_\mathit{IC}$ pattern  $(\cap >s_2, \cap \leq s_2, \subseteq >s_1)$ is  interpreted as $(\Delta \supseteq [s_2, s_2+\epsilon],\Delta \subseteq ]s_1, s_3])$  (node 19 in Figure\ref{figLIII-s1s2}) and  only occurs in object $o_4 $. 
\end{example}

Finally, when considering $L_\mathit{IC}^2$ displayed Figure \ref{figLIII-s1s2} it is interesting to note that  
all singletons $\{\Delta_i\}$  are represented as lowest nodes of the lattice above $\bot$:
%

\begin{proposition}
All distinguishable intervals when considering $T=\{1 \dots k\}$ are represented as singletons in $L_\mathit{IC}^k$.
\end{proposition}

\section{Experiments on  uncertain data}\label{experiments}
In order to experiment our interval pattern languages 
we transformed the well known  Iris dataset by replacing numerical  values by intervals containing the real value. 
The transformation protocol  was previously used on this dataset in \cite{Gullo:2008aa}.
In the Iris dataset there  are 150 iris flower each described by 4 variables representing petal length ($P_l$) and  width ($P_w$),  together with sepal length ($S_l$) and width ($S_w$). The flowers are partitioned into 3 classes corresponding to three species. The protocol to transform numerical values to intervals is as follows:
 For each variable  we consider  two bounds $\delta_M$ and  thresholds $\{s_1, \dots s_4\} $ obtained by considering 5 equal-width bins. Then, the value $y$ of each variable is replaced by an interval $\Delta_y$ by
	 i)  Drawing a  size $w$ uniformly between $0$ and  $(\delta_M- \delta_m)/2$ and ii) 
	 Defining  $\Delta_y=] y_l, y_r]$ where $y_l=\mathrm{max}(\delta_m,y-w)$ and $y_r=\mathrm{min}
	 (\delta_M,y+w) $.
In what follows we firs look at how well interval patterns represent the original iris classes, then we compare clustering results using the various interval pattern languages.
 \subsection{$L_\mathit{I}$, $L_\mathit{C}$ and $L_\mathit{IC}$ closed patterns and distance to  classification }
We enumerate  closed patterns with support at least $20$ in various settings:
\begin{enumerate}
\item   $L_\mathit{Num}^4$ in which for  variable $Y$ we consider   $y> s_i$ and  $y\leq  s_i$ constraints.
\item Transformed dataset with, for each interval variable,   pattern language  $L_\mathit{C}^4$ 
\item Transformed dataset with, for each interval variable,     pattern language $L_\mathit{I}^4$ 
\item  Transformed dataset with, for each interval variable,    pattern language  $L_\mathit{IC}^4$ 
\end{enumerate}

The results are summarized Table \ref{tableIris} which reports in the four settings the \emph{number of closed patterns}  with support at least 20,  the \emph{covering} of the set of closed patterns, i.e. the number of flowers belonging to at least one of these support sets, the \emph{average  size of these support sets} and the \emph{distance between the set of support sets  and the original partition} of the 150 flowers in 3 classes (see below). We also report the \emph{distance between the  set of  support sets and a random partition} in three classes of 50 flowers\footnotemark in order to evaluate significance of the difference in distances of the support sets to the original partition.
\footnotetext{Obtained by random exchanges between elements of classes in the original partition}
 The distance between a set of support sets $S$ and a partition $C$ is defined as 
$d_\mathit{s}(S,C)=(1/|C|) \sum_{c\in C} min_{s\in S} d_J(c,s)$  
 \mbox{      where }  $ d_J(s,c)$ is the Jaccard distance between subsets $c$ and $s$. 
Informally this distance
consider each class $c$ in $C$ and looks for the closest subset $s$  in $S$.
 Whenever all classes in $C$ are also found as subsets in $S$ this distance is null 
  and we obtain the  true classification by  identifying one  pattern for each class.

\begin{table}\begin{center}
$
\begin{array}{cccccc}
 \text{Language} & \text{$\#$Patterns} & \text{Covering}&\overline{\text{Support}} &\text{Dist. to Classes} &\text{Dist. to Random} \\
 \hline
 L_\mathit{Num}^4	& 1948	& 150		& 41.213		& 0.110707	&	0.65\\
   L_\mathit{C}^4	& 84		&150			& 39.3214		& 0.297969	&	0.66 \\
  L_\mathit{I}^4		& 46985	&150			& 50.7512		& 0.209509	&	0.63\\
   L_\mathit{IC}^4 	& 121181 	&150 		& 41.4479 	& 0.166609	&	0.63
\end{array}
$
\end{center}
\caption{Closed patterns in from the iris  dataset using  $L_\mathit{Num}$
$L_\mathit{C}^4$,  $L_\mathit{I}^4 $ and $ L_\mathit{IC}^4$ }\label{tableIris}
\end{table} 

As expected we  observe that  the number of closed patterns in  $L_\mathit{C}^4$ is very small compared to those obtained with $ L_\mathit{Num}^4$, while the number of closed patterns in  $L_\mathit{I}^4$ and $L_\mathit{IC}^4$ is much larger. 
Regarding the distance of the sets of support sets to the original partition and to random partitions, we first observe for all languages large distances (beyond 0.6)  to random partitions. On the contrary, distances to the original partition display differences between languages. The smallest distance (0.11) is as expected obtained from  $L_\mathit{Num}$.
Adding random uncertainty to the variables  decreases their ability to retrieve the original classes, which translates into a higher distance between the original partition and the sets of support sets. The worst result ($\approx 0.3$) is obtained with $L_\mathit{C}^4$  that still  is much  lower than in the random case. The distance obtained with $ L_\mathit{I}^4$ is lower ($\approx 0.21$) and the distance obtained with $L_\mathit{IC}^4$ is even lower  ($\approx 0.17$) closer to the distance obtained with $L_\mathit{Num}$.
 We also observe a total covering whatever is the language.
 
 \subsection{Clustering  with interval pattern languages}
 
 We have experimented  clustering of these modified datasets. For these tasks we added a $L_\mathit{Ori}$  line referring to the original description of objects as a vector of 4 numerical values.  We report Table \ref{tableIrisClassificationAndClustering} the overall F-measure (see \cite{Gullo:2008aa})  of the clustering result when compared to the original   three classes partition.
Clustering is performed using the k-medoids algorithm ($k=3$). We averaged   the  overall F-measure on 200 trials.
\begin{table}\begin{center}
$
\begin{array}{cccccc}
 \text{Language} & \text{F-measure} & \text{std} 	\\
 \hline
 L_\mathit{Ori}		& 0.9	00		&\pm 0.001	\\
 L_\mathit{Num}^4	& 0.860		&\pm 0.036	\\
   L_\mathit{C}^4	& 0.814		&\pm 0.059	\\
  L_\mathit{I}^4		& 0.856		&\pm 0.041	\\
   L_\mathit{IC}^4 	& 0.841		&\pm 0.039	\\
\end{array}
$
\end{center}
\caption{
Average overall F-measure of 3-medoids clustering on 200 trials. 
}\label{tableIrisClassificationAndClustering}
\end{table} 
Table \ref{tableIrisClassificationAndClustering} gives answers on  questions  regarding clustering:
\begin{itemize}
\item
 $ L_\mathit{Ori}$	vs $ L_\mathit{Num}$: 
Clustering results display a clear F-measure loss  ($0.90$ vs $0.86$) when compared  to the reference three classes partition. 
\item $ L_\mathit{Num}^4$	vs $ L_\mathit{C}^4, L_\mathit{I}^4, L_\mathit{IC}^4$:
What is the information loss resulting from adding uncertainty to scaled values?   
As a result of performing Student t-tests $ L_\mathit{Num}^4$  F-measure ($0.86$)  differs significantly (with p-value $<10^{-4}$) from  $L_\mathit{C}^4$ ($0.814$) and from $L_\mathit{IC}^4$  ($0.841$) but does not differ significantly from  the $L_\mathit{I}^4$ ($0.856$).  Clearly  $ L_\mathit{C}^4$ results in a F-measure loss while   $ L_\mathit{I}^4$ does not, and  $ L_\mathit{IC}^4$
display  worse but close to  $ L_\mathit{I}^4$. 

\end{itemize} 

%

We have   compared our clustering  results on the iris dataset to those  presented in \cite{Gullo:2008aa}.
 The   k-medoid clustering  algorithm adapted to  uncertain data 
resulted in   a 0.84 F-measure value  close to the values we obtained using  $L_\mathit{I}^4$ (0.856)  and $L_\mathit{IC} ^4$ (0.841). 
 This suggests   that handling intervals using relevant languages allows using standard clustering algorithms with results comparable to those obtained by methods specifically designed to handle intervals.

%
%




 \section{Patterns on distributional data}\label{distributionalData}
 
 An interval variable $\Delta$ may come from various kind of data,  among which  those where $\Delta$  represents uncertainty  on some numerical variable $Z$.
   We consider then $Z$  as a random variable whose cumulative distribution function (\textit{cdf}) $F$, defined as $F(z)=p(Z\leq z)$, depends on the object $o$. 

\subsubsection{$L_{I^\alpha}$}
 We  define then the language $L_I^\alpha$, with $0<\alpha< 0.5$  as built on the following  atomic patterns  
 for all thresholds  $s\in S$:
\begin{eqnarray}
	 F(s) > \alpha		  & , &F(s) <1-\alpha \label{inside}
\end{eqnarray}

\subsubsection{$L_{\mathit{IC^\alpha}}$}
We add to the two atomic patterns of $L_{I^\alpha}$  their negations
\begin{eqnarray}
	 F(s) \leq \alpha		  & , &F(s) \geq 1-\alpha \label{outside}
\end{eqnarray}
 
Let us  then  define the interval $\Delta_\alpha$ with $F$ values between $\alpha$ and $1-\alpha$, i.e. $ \Delta_\alpha	=\{y, \mid F(y)>\alpha, \ \  F(y)\leq  1-\alpha  \}$  and express these distributional  constraints as interval constraints:
   \begin{eqnarray}
	  \Delta_\alpha \ \cap \ ]\delta_m, s] \not =\emptyset					&\equiv  &		 F(s) > \alpha		 \label{Fgta}\\
  	   \Delta_\alpha \ \cap \ ]s, \delta_M] \not =\emptyset 			& \equiv & 			 F(s)  < 1-  \alpha 	\label{Ggta}\\
	        \Delta_\alpha \ \subseteq \ ]s, \delta_M] 				& \equiv &		F(s) \leq \alpha 	 	\label{Flea}\\
       \Delta_\alpha \ \subseteq \ ]\delta_m, s [\not =\emptyset				& \equiv &			F(s) \geq  1-\alpha \label{Glea}
         \end{eqnarray}


\subsubsection{Continuous distributions}
As  $F$ is continuous and strictly increasing on $[q_0,q_1]$, we have that  for any  $v \in[0,1]$,  $q^v=F^-1(v)$ exists.With $\alpha\in]0,0.5[$ we have:
\begin{itemize}
\item $ F(s) > \alpha$ rewrites as $s> q^\alpha$, $ F(s) < 1-\alpha$ rewrite as $s<q^{1-\alpha}$
\item $ F(s) \leq \alpha$ rewrites as $s\leq q^\alpha$, $ F(s)> 1-\alpha$ rewrites as $s\geq q^{1-\alpha}$
\end{itemize}
With $L_{I^\alpha}$ we may  express for any object $o$ whether $s$ is  on the right side of $q^\alpha$ and whether it is on the left side of $q^{1-\alpha}$. We may then express whether $s$ belongs to the interval $I^\alpha=]q^\alpha, q^{1-\alpha}[$ such that  $p(Z\in I^\alpha)=p(Z\in \Delta_\alpha)=F(1-\alpha)-F(\alpha) = 1-2\alpha$.  
With the full $L_{IC^\alpha}$ language, we may also express whether $s$ is on the left side of $q^\alpha$ and whether $s$ is on the right side of $q^{1-\alpha}$.
Now a distribution is often represented by  its $n$-quantiles $q_i$ such that $F(q_i)= i/n$. 
When $\alpha=1/n$, $q^\alpha$ and $q^{1-\alpha}$ represent the first and last $n$-quantile of the distribution.

 
 \begin{example}
  Consider 10-quantiles, i.e. deciles,  and an infinite domain $D$ with  thresholds $T=\{-0.5,2.5,4.5,6.5\}$.
 
Consider an object $o_1$  whose $Z$ variable follows a normal distribution $\mathcal{N}(2,1)$.  The  bounds $q_1$ and $q_9$  as well as the threshold $s_2=2.5$ are reported Figure\ref{Fig-DensityAndRepartition}.
We have $ q_1< s_2< q_9$ which means that   pattern
$1/10<F(s_2) <9/10$ occurs in $o_1$.
%
 
 \begin{figure}[h]
    \centering

      \includegraphics[width=0.30\textwidth]{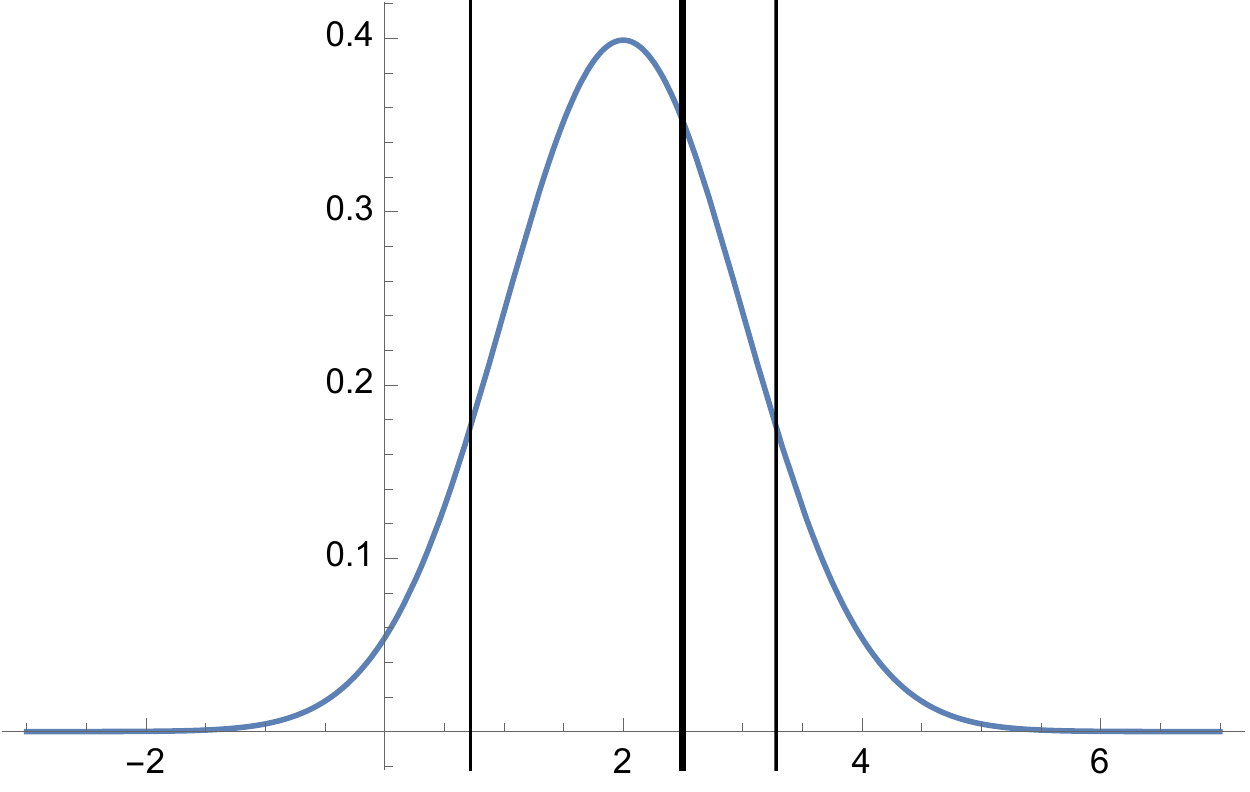} \ \ \ \ \ \ \ 
          \includegraphics[width=0.30\textwidth]{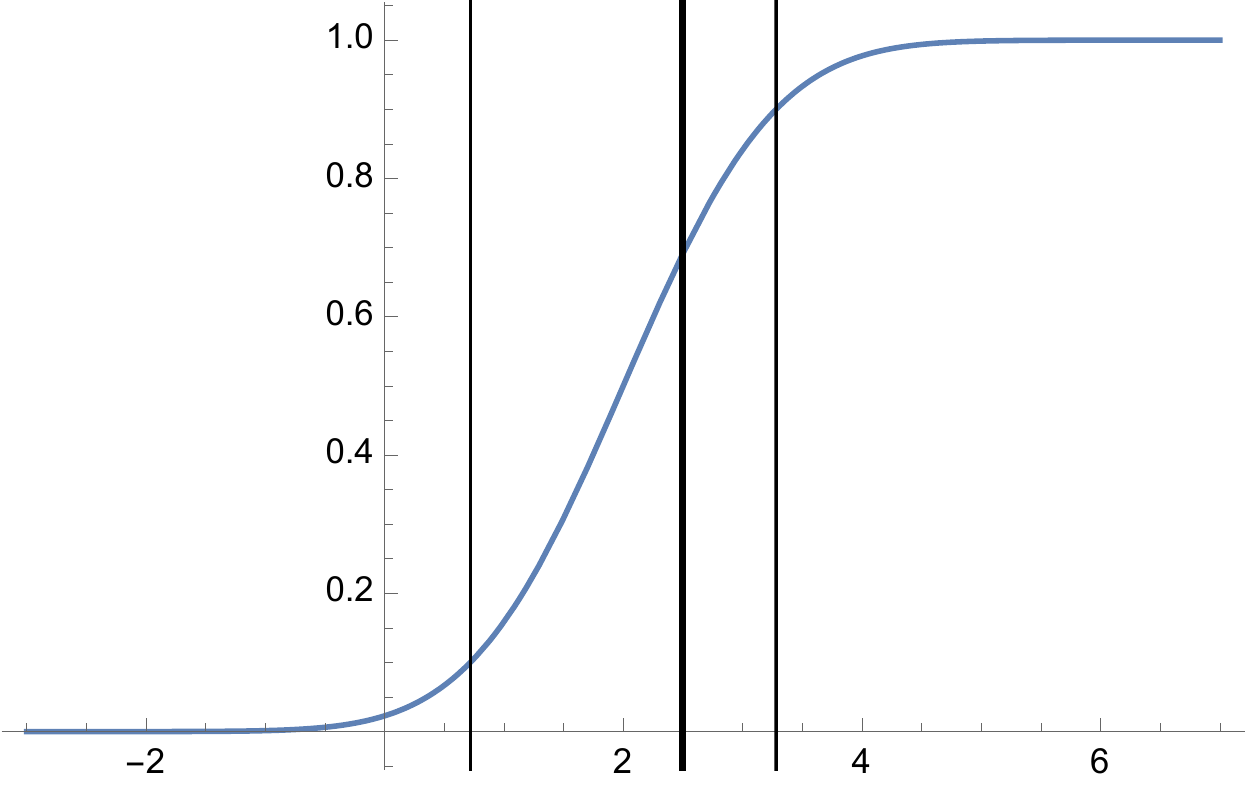}
    \caption{Density (left) and repartition function(right) of $\mathcal{N}(2,1)$ distribution. 
    The vertical bars figure  $q_1$ and  $q_9$  while the  thick vertical bar represents $s_2=2.5$.
    }
  \label{Fig-DensityAndRepartition}
\end{figure}

  \end{example}
  \subsubsection{Discrete distributions }
Whenever $F$ is discrete, the $Z$ values belong to some countable subset $C$ of $\mathcal{R}$. 
To express patterns  with respect to the $s$ value  we need to define $q^\alpha$ and $q^{1-\alpha}$ for any $\alpha$.
 To match our definitions we define them  as $q^\alpha= \mathrm{max}\{c \in C \mid F(x)\leq \alpha\}$ and $q^{1-\alpha}= \mathrm{min}\{c \in C \mid F(x)\geq 1-\alpha\}$.

\begin{example}
Consider the discrete uniform distribution on  $C=1..4$ and  $\alpha=1/5$.
\begin{itemize}

\item$F(s) > 1/5$  happens whenever $s> 0$.  So $q^{1/5}=0$ and we have indeed $ \mathrm{max}\{c \in C \mid F(x)\leq 1/5\}=0$
\item$F(s) < 4/5$ happens whenever $s < 4$. So $q^{4/5}=4$ and we have indeed $ \mathrm{min}\{c \in C \mid F(x)\geq 4/5\}=4$
\end{itemize}
\end{example}
%
  
  \subsection{Simulation of uncertainty and classification experiments}
In our previous experiments in Section \ref{experiments}, for each variable $Z$ the true value  $z$ in   object $o$  were in most cases the center of the interval representing $Z$ uncertainty in $o$. To investigate the effect of uncertainty on classification we need to introduce uncertainty in  a more realistic way. We propose  the following scenario:  
  
  \begin{enumerate}
  \item The experimenter measures $Z$  with a device returning  a value following a Normal distribution $F_z$  centered on the unknown true value $z$ with a standard deviation $\sigma$ which only depends on the device.
    \item The device returns the value  $v$. The experimenter then  represents the  $Z$ value by a Normal distribution $F_v$ centered around $v$ with  standard deviation  $\sigma$.  
  
  \end{enumerate} 
  We benefit then from the following proposition:
  \begin{proposition}
  If  the distribution is symmetric with median value $m$, i.e. $F(m-z)=1-F(m+z)$, we have for any $\alpha \in]0,0.5]$
  \begin{equation}
  \alpha < F_z(v) < 1-\alpha \mbox{ if and only if }   \alpha < F_v(z) < 1-\alpha
  \end{equation}
  \end{proposition}
 This means that if the observed value $v$ is within a range around  the true value $z$ of  probability $1-2 \alpha$,  $z$ also is within a range of same probability around $v$.
 
 In our experiments we apply the same treatment to the four variables:
 \begin{itemize}
 \item We  draw a value of $\sigma$ for each object, and then draw  a value $v$  following $F_z$. We obtain a new dataset \emph{IrisO} in each object  of which $z$ is replaced by   $v$. We then build  \emph{IrisOD}   each object of which   is represented by the distribution $F_v$. 
 \item We encode  \emph{IrisO} and  \emph{IrisOD} with  threshold sets $T^4$   and $T^8$.
\emph{IrisO} is encoded into  \emph{IrisNumO}. \emph{IrisOD}  is  encoded  into  \emph{IrisOD$\alpha$} using $L_I^{0.1}$ and into  \emph{IrisOD}$_{0.1,0.25}$ using  
   both atoms from $L_I^{0.1}$ and from $L_I^{0.25}$. 
 \end{itemize}%

For each object $o$, the standard deviation $\sigma$  of $F_z$ is drawn as follows.
Let $z_\mathit{min}, z_\mathit{max}$  be the extrema  of  $Z$ on the dataset, we assign them as first and last deciles of a  Normal   distribution $\mathcal{N}((z_\mathit{min}+z_\mathit{max})/2,\sigma_\mathit{max})$.
Then,  we  draw 
$\sigma_o$   from $]0,\sigma_\mathit{max}/r]$ where $r$ is a reduction factor fixed at $r=1.75$.
%

We ran then classification experiments, using a Random Forest classifier on these various datasets. Table \ref{tableIris1point75ClassificationAndClustering} shows that for both 4 and 8 thresholds the distributional encodings $L_\mathit{OD_{{0.1}\textit{-}\mathit{0.25}}}$ display a better accuracy than direct value $v$ encoding. The 8 thresholds $L_\mathit{OD_{{0.1}\textit{-}\mathit{0.25}}}$ accuracy ($0.906$)  is even better than the one of raw $v$ value ($0.886$).

       \begin{table}
 \begin{minipage}[c]{0.55\textwidth}
\begin{center}
$
\begin{array}{cccccccccccc}
 \text{Language.}				&\text{Acc.} & \text{std} & \text{Language.}				&\text{Acc.} & \text{std} \\
 \hline
 L_\textit{O}					&0.886	&\pm 0.015				& 					&		&				\\
 L_\mathit{NumO}^4				&0.865	&\pm 0.018				&  L_\mathit{NumO}^8	&0.881	&\pm 0.012			\\

 L_\mathit{OD_{0.1}}^4			&0.883	&\pm 0.012				& L_\mathit{OD_{0.1}}^8			&0.881	&\pm 0.013			\\
L_\mathit{OD_{{0.1}\textit{-}\mathit{0.25}}}^4		&0.884	&\pm 0.013 	& L_\mathit{OD_{{0.1}\textit{-}\mathit{0.25}}}^8		&0.906	&\pm 0.013			\\
 \end{array}
$
\end{center}
  \end{minipage}\hfill
  \begin{minipage}[c]{0.40\textwidth}
  \begin{center}
  \caption{10-10 cross-validation classification accuracies  on the  simulated  datasets. 
}\label{tableIris1point75ClassificationAndClustering}
\end{center}
  \end{minipage}\hfill
\end{table}

\section{Conclusion}
We have investigated the interval pattern language $L_\mathit{I}$, which is based on  intersection-based constraints and compared  in term of expressivity, and experimentally  $L_\mathit{I}$ to  the inclusion-based pattern language  $L_\mathit{C}$ and to 
$L_\mathit{IC}$  that combines intersection based  and inclusion based constraints.
All these languages are lattices ordered by specificity, adequate   for  FCA and closed pattern mining and 
may be implemented in the standard  closed itemset setting. As shown in the various experiments $L_\mathit{I}$ results in much more closed patterns, with a wide range from very specific to very general patterns. In contrast,  the standard inclusion-based approach implemented in $L_\mathit{C}$ tends to leads to overgeneral patterns even with small supports. 
We have also applied the approach to distributional data, by representing a distribution by a small set of intervals, each corresponding to first and last quantiles. Our first experiments, realistically simulating data uncertainty, show some increase in accuracy with respect to simply encoding the simulated value. Still these results are preliminary and deserve more investigations.  


\bibliographystyle{unsrt}  
\bibliography{references} 


\end{document}